\documentclass[12pt]{article}
\usepackage{amssymb}                         

\newtheorem{theorem}{Theorem}
\newtheorem{lemma}{Lemma}
\newtheorem{corollary}{Corollary}
\newtheorem{proposition}{Proposition}
\newtheorem{definition}{Definition}

\newcommand {\Mybox}{\Box}

\def\x{{\bf x}}
\def\y{{\bf y}}

\def\N{\mathbb N}

\date{}
\title{\large{\textbf{Non-Elitist Genetic Algorithm as a Local Search Method}}}
\author{\small{Anton V. Eremeev}}
\begin{document}
\maketitle
\begin{center}

Omsk Branch of Sobolev Institute of Mathematics SB RAS\\
13, Pevstov str., 644099, Omsk, Russia\\
e-mail: {\tt eremeev@ofim.oscsbras.ru}\\

\end{center}

{\bf Abstract.} Sufficient conditions are found under which the
iterated non-elitist genetic algorithm with tournament selection
first visits a local optimum in polynomially bounded time on
average. It is shown that these conditions are satisfied on a
class of problems with guaranteed local optima~(GLO)
if appropriate parameters of the algorithm are chosen.\\


\section*{\large{\textbf{Introduction}}}

The genetic algorithm (GA) proposed by J.~Holland~\cite{Holl75} is
a randomized heuristic search method, based on analogy with the
genetic mechanisms observed in nature and employing a population
of tentative solutions. Different modifications of GA are widely
used in the areas of operations research pattern recognition,
artificial intelligence etc. (see e.g.~\cite{Reeves,Yank99}).
Despite of numerous experimental investigations of these
algorithms, their theoretical analysis is still at an early
stage~\cite{BSW}.

In this paper, the genetic algorithms are studied from the
prospective of local search for combinatorial optimization
problems, and the NP~optimization problems in
particular~\cite{AP95}. The major attention is payed to
identification of the situations where the GA finds a local
optimum in polynomially bounded time on average. Here and below we
assume that the randomness is generated only by the randomized
operators of selection, crossover and mutation within the GA. In
what follows, we call a value {\em polynomially bounded}, if there
exists a polynomial in the length of the problem input, which
bounds the value from above. Throughout the paper we use the terms
{\em efficient algorithm} or {\em polynomial-time algorithm} for
an algorithm with polynomially bonded running time. A problem
which is solved by such an algorithm is {\em polynomially
solvable.}

This study is motivated by the fact that the GAs are often
considered to be the local search methods (see
e.g.~\cite{AartsLenstra,Krish,RR02}). Therefore a topical question
is: In what circumstances the GA efficiency is due to its
similarity with the local search?

\section{Standard definitions and algorithm description}\label{sec:defs}

\paragraph{NP Optimization Problems.}

In what follows, the standard definition of an NP~optimization
problem is used (see e.g.~\cite{ACGKMP}). By $\{0,1\}^*$ we denote
the set of all strings with symbols from~$\{0,1\}$ and arbitrary
string length. For a string~$S\in \{0,1\}^*$, the symbol~$|S|$
will denote its length. In what follows $\N$ denotes the set of
positive integers, and given a string~$S\in \{0,1\}^*$, the
symbol~$|S|$ denotes the length of the string~$S$.

\begin{definition}\label{def:NPO} An NP~optimization problem $\Pi$
is a triple ${\Pi=(\mbox{\rm Inst},\mbox{\rm Sol}(I),f_I)}$, where
$\mbox{\rm Inst} \subseteq \{0,1\}^*$
is the set of instances of~$\Pi$ and:\\

1. The relation $I \in \mbox{\rm Inst}$ is computable in polynomial time.\\

2. Given an instance $I \in \mbox{\rm Inst}$, $\mbox{\rm
Sol}(I)\subseteq \{0,1\}^{n(I)}$ is the set of feasible solutions
of~$I$, where~$n(I)$ stands for the dimension of the search
space~$\{0,1\}^{n(I)}$. Given $I \in \mbox{\rm Inst}$ and $\x \in
\{0,1\}^{n(I)}$, the decision whether $\x\in \mbox{\rm Sol}(I)$
may be done in polynomial time, and $n(I) \le \mbox{\rm
poly}(|I|)$ for some polynomial~$\mbox{\rm
poly}$.\\

3. Given an instance~$I \in \mbox{\rm Inst}$,  $f_I: \mbox{\rm
Sol}(I) \to {\N}$ is the objective function (computable in
polynomial time) to be maximized if $\ \Pi$ is an NP~maximization
problem or to be minimized if $\ \Pi$ is an NP~minimization
problem.
\end{definition}

Without loss of generality we will consider in our analysis only
the maximization problems. The results will hold for the
minimization problems as well. The symbol of problem instance~$I$
may often be skipped in the notation, when it is clear what
instance is meant from the context.

\begin{definition}\label{PolyBoundedProblem}
A combinatorial optimization problem is {\em polynomially
bounded}, if there exists a polynomial in~$|I|$, which bounds the
objective values $f_I({\bf x})$, ${\bf x} \in {\rm Sol}(I)$ from
above.
\end{definition}

An algorithm for an NP~maximization problem~$\Pi=(\mbox{\rm
Inst},\mbox{\rm Sol}(I),f_I)$ has a {\em guaran\-teed
approxi\-mation ratio}~$\rho$, $\rho \ge 1$, if for any
instance~$I\in {\rm Inst}, \ {\rm Sol}(I) \ne \emptyset,$ it
delivers a feasible solution~$\x$, such that $f_I(\x) \ge
\max\{f_I(\x) | \x \in {\rm Sol}(I)\}/\rho$.

\paragraph{Neighborhoods and local optima.}

Let a neighborhood ${\mathcal N}_I({\bf y})\subseteq \mbox{\rm
Sol}(I)$ be defined for every~${\bf y}\in \mbox{\rm Sol}(I)$. The
mapping ${\mathcal N}_I: {\rm Sol}(I) \to 2^{{\rm Sol}(I)}$ is
called the {\em neighborhood mapping}. This mapping is supposed to
be efficiently computable (see e.g.~\cite{AP95}).

\begin{definition} If the inequality $f_I({\bf y})\leq f_I({\bf x})$
holds for all neighbors~$\y \in {\mathcal N}_I({\bf x})$ of a
solution~${{\bf x} \in \mbox{\rm Sol}(I)}$, then~${\bf x}$ is
called a local optimum w.r.t. the neighborhood mapping~${\mathcal
N}_I$.
\end{definition}

Suppose $D(\cdot,\cdot)$ is a metric on $\mbox{\rm Sol}(I)$. The
neighborhood mapping
$$
{\mathcal N}_I({\bf x})=\{{\bf y}: D({\bf x},{\bf y}) \leq R\}, \
\ {\bf x} \in \mbox{\rm Sol}(I),
$$
is called a neighborhood mapping of radius~$R$ defined by
metric~$D(\cdot,\cdot)$.

A local search method starts from some feasible solution~$\y_0$.
Each iteration of the algorithm consists in moving from the
current solution to a new solution in its neighborhood, such that
the value of objective function is increased. The way to choose an
improving neighbor, if there are several of them, will not matter
in this paper. The algorithm continues until it will reach a local
optimum.

\paragraph{Genetic Algorithms.}

The simple GA proposed in~\cite{Holl75} has been intensively
studied and exploited over four decades. A plenty of variants of
GA have been developed since publication of the simple GA, sharing
the basic ideas, but using different population management
strategies, selection, crossover and mutation
operators~\cite{RR02}.

The GA operates with populations~$X^t=({\bf x}^{1t},\dots,{\bf
x}^{\lambda t})$, \ $t=1,2,\dots,$ which consist of~$\lambda$ {\it
genotypes}. In terms of the present paper the genotypes are
strings from~$B=\{0,1\}^n$. For convenience we assume that the
number of genotypes~$\lambda$ is even.

In a {\em selection} operator~${\mbox{Sel}: B^{\lambda} \to
\{1,\dots,\lambda\}}$, each parent is independently drawn from the
previous population~$X^t$ where each individual in~$X^t$ is
assigned a selection probability depending on its {\em
fitness}~$\Phi({\bf x})$. Below we assume the following natural
form of the fitness function:

\begin{itemize}

\item if $\x\in {\rm Sol}$ then
$$\Phi({\bf x}) = f({\bf x});$$

\item if ${\bf x}\not \in {\rm Sol}$ then its fitness is
defined by some penalty function, such that
$$
\Phi({\bf x}) < \min_{\y \in {\rm Sol}} f(\y).
$$
\end{itemize}

In this paper we consider the tournament selection
operator~\cite{Gold90}: draw~$k$ individuals uniformly at random
from~$X^t$ (with replacement) and choose the best of them as a
parent.

A pair of offspring genotypes is created using the randomized
operators of crossover~${\mbox{Cross}: B \times B \to B \times B}$
and mutation ${\mbox{Mut}:B \to B}$.
%
%
%
%
In general, we assume that operators $\mbox{Cross}(\x,\y)$ and
$\mbox{Mut}(\x)$ are efficiently computable randomized routines.
We also assume that there exists a positive constant~$\varepsilon$
which does not depend on~$I$, such that the fitness of at least
one of the genotypes resulting from crossover $({\bf x}',{\bf
y}')=\mbox{Cross}({\bf x},{\bf y})$ is not less than the fitness
of the parents ${\bf x},{\bf y} \in B$ with probability at
least~$\varepsilon$, i.e.
\begin{equation}\label{eps_cross}
{\bf P}\big\{\max\{\Phi({\bf x}'),\Phi({\bf y}')\} \ge
\max\{\Phi({\bf x}),\Phi({\bf y})\}\big\} \ge \varepsilon
\end{equation}
for any ${\bf x},{\bf y} \in B$.

When a population~$X^{t+1}$ of~$\lambda$ offspring is constructed,
the GA proceeds to the next iteration~$t+1$. An initial
population~$X^{0}$ is generated randomly. One of the ways of
initialization consists in independent choice of all bits in
genotypes.

To simplify the notation below, $\mathcal GA$ will always denote
the non-elitist genetic algorithm with the following
outline.\\

{\bf Algorithm $\mathcal GA$}\\
\vspace{-0.5em}

\noindent Generate the initial population~$X^0$, assign $t:=0.$\\
 {\bf While} termination condition is not met {\bf do:}\\
$\mbox{\hspace{2em}}$ {\bf Iteration $t+1$}:\\
$\mbox{\hspace{2em}}$ {\bf For $j$ from 1 to ${\lambda}/2$ do:} \\
$\mbox{\hspace{4em}}$   Tournament selection: $\x:=\x^{{\rm Sel}(X^t),t}$, $\y:=\x^{{\rm Sel}(X^t),t}.$ \\
$\mbox{\hspace{4em}}$   Mutation: $\x' := \mbox{Mut}(\x), \ \ \y':=\mbox{Mut}(\y).$\\
$\mbox{\hspace{4em}}$   Crossover: $(\x^{2j-1,t+1},\x^{2j,t+1}):=\mbox{Cross}(\x',\y').$\\
$\mbox{\hspace{2em}}$ {\bf End for.}\\
$\mbox{\hspace{2em}}$ $t:=t+1.$\\
 {\bf End while.}\\

The population size~$\lambda$ and tournament size~$k$, in general
may depend on problem instance~$I$. The termination condition may
be required to stop a genetic algorithm when a solution of
sufficient quality is obtained or the computing time is limited,
or because the population is "trapped" in some unpromising area
and it is preferable to restart the search. In theoretical
analysis of the $\mathcal GA$ it is often assumed that the
termination condition is never met. In order to incorporate the
possibility of restarting the search, we will also consider
the iterated~$\mathcal GA$, which has the following outline.\\

{\bf Algorithm iterated $\mathcal GA$}\\
\vspace{-0.5em}

\noindent {\bf Repeat:}\\
$\mbox{\hspace{2em}}$ Generate the initial population~$X^0$, assign $t:=0.$\\
$\mbox{\hspace{2em}}$ {\bf While} termination condition $t>t_{\max}$ is not met {\bf do:}\\
$\mbox{\hspace{4em}}$ Iteration $t+1$:\\
$\mbox{\hspace{4em}}$ {\bf For $j$ from 1 to ${\lambda}/2$ do:} \\
$\mbox{\hspace{6em}}$   Tournament selection: $\x:=\x^{{\rm Sel}(X^t),t}$, $\y:=\x^{{\rm Sel}(X^t),t}.$ \\
$\mbox{\hspace{6em}}$   Mutation: $\x' := \mbox{Mut}(\x), \ \ \y':=\mbox{Mut}(\y).$\\
$\mbox{\hspace{6em}}$   Crossover: $(\x^{2j-1,t+1},\x^{2j,t+1}):=\mbox{Cross}(\x',\y').$\\
$\mbox{\hspace{4em}}$ {\bf End for.}\\
$\mbox{\hspace{4em}}$ $t:=t+1.$\\
$\mbox{\hspace{2em}}$ {\bf End while.}\\
{\bf Until false.}\\


\paragraph{Examples of mutation and crossover operators.}

As examples of crossover and mutation we can consider the
well-known operators of bitwise mutation~$\mbox{Mut}^*$ and
single-point crossover~$\mbox{Cross}^*$ from the simple
GA~\cite{Holl75}.

The crossover operator computes $({\bf x}',{\bf
y}')=\mbox{Cross}^*({\bf x},{\bf y})$, given $ {\bf
x}=(x_1,...,x_n ),$ ${\bf y}=(y_1,..., y_n),$ such that with
probability~$P_{\rm c}$,
$$
{\bf x}'=(x_1,...,x_{\chi}, y_{\chi+1},...,y_n ), \ \  {\bf
y}'=(y_1,..., y_{\chi}, x_{\chi+1},..., x_n),
$$
where the random number~$\chi$ is chosen uniformly from~1
to~$n-1$. With probability ${1-P_{\rm c}}$ both parent individuals
are copied without any changes, i.e. ${\bf x}'={\bf x}, \ {\bf
y}'={\bf y}$.

Condition~(\ref{eps_cross}) is fulfilled for the single-point
crossover with $\varepsilon = 1-P_{\rm c}$, if $P_{\rm c}<1$ is a
constant. Condition~(\ref{eps_cross}) would also be satisfied with
$\varepsilon = 1$, if an optimized crossover operator was used
(see e.g.,~\cite{BN98,ErECJ08}).

The bitwise mutation operator~$\mbox{Mut}^*$ computes a
genotype~${\bf x}'=\mbox{Mut}^*({\bf x})$, where independently of
other bits, each bit~$x'_i, \ i=1,\dots,n$, is assigned a
value~$1-x_i$ with probability~$P_{\rm m}$ and with
probability~$1-P_{\rm m}$ it keeps the value~$x_i$.



\section{Expected Hitting Time of a Local Optimum}
\label{sec:runtime}

Suppose an NP~maximization problem~$\Pi=({\rm Inst},\mbox{\rm
Sol}(I),f_I)$ is given and a neighborhood mapping~$\mathcal{N}_I$
is defined. Let~$m$ denote the number of all non-optimal values of
objective function~$f$, i.e. $m=|\{f \ | \ f=f({\bf x}),\ {\bf x}
\in \mbox{\rm Sol} \}|-1$. Then starting from any feasible
solution the local search method finds a local optimum within at
most~$m$ steps. Let us compare this process to the computation of
a~$\mathcal GA$.

Let~$s$ be a lower bound on the probability that the mutation
operator transforms a given solution~$\x$ into a specific
neighbor~${\bf y} \in {\mathcal N} ({\bf x})$, i.e.
$$
s \le \min_{{\bf x} \in {\rm Sol}, \ {\bf y} \in {\mathcal N}
({\bf x})}{\bf P}\{\mbox{Mut}({\bf x})={\bf y}\}.
$$
The greater the value~$s$, the more consistent is the mutation
with the neighborhood mapping~${\mathcal N}$. Let the size of
population~$\lambda$, the tournament size~$k$ and the bound~$s$ be
considered as functions of the input data~$I$. The symbol~$e$
denotes the base of the natural logarithm.

\begin{lemma}\label{lemma_GA_LS}
If $X^0$ contains a feasible solution, $k \ge r\lambda$, $r>0$, $m
> 1$, $s>0$ and
\begin{equation}\label{N_condition}
\lambda \ge \frac{2(1+ \ln m)}{s\varepsilon(1-1/e^{2r})},
\end{equation}
then the $\mathcal GA$ visits a local optimum until iteration~$m$
with probability at least~$1/e$.
\end{lemma}

{\bf Proof.} Note that in the initial population, the individual
of greatest fitness is a feasible solution. Let an
event~$E_j^{t+1}$, $j=1,\dots,\lambda/2$, consist in fulfilment of
the following three conditions:
\begin{enumerate}
\item An individual~${\bf x}^t_*$ of greatest fitness in
population~$X^t$ is selected at least once when the $j$-th pair of
offspring is computed;
 \item Mutation operator applied to~${\bf x}^t_*$
performs the best improving move within the
neighborhood~${\mathcal N}({\bf x}^t_*)$, i.e.
$\Phi(\mbox{Mut}({\bf x}^t_*))=\max_{{\bf y} \in {\mathcal N}({\bf
x}^t_*)} \Phi({\bf y})$.
 \item When the crossover operator is
applied for computing the $j$-th pair of offsprings, at least one
of its outputs has the fitness not less than~$\max_{{\bf y} \in
{\mathcal N}({\bf x}^t_*)} \Phi({\bf y})$;
\end{enumerate}

Let~$p$ denote the probability of union of the events~$E_j^{t+1},
\ j=1,\dots,\lambda/2$. In what follows we construct a lower
bound~$\ell \le p$, which holds for any population~$X^t$
containing a feasible solution. According to the outline of
the~$\mathcal GA$, ${\bf P}\{E_1^{t+1}\}=\dots ={\bf
P}\{E_{\lambda/2}^{t+1}\}$. Let us denote this probability by~$q$.
Given a population~$X^t$, the events $E_j^{t+1}, \
j=1,\dots,\lambda/2,$ are independent, so $p \ge
1-(1-q)^{\lambda/2} \ge 1-e^{-q \lambda/2}$. Now~$q$ may be
bounded from below:
$$
q \ge s \varepsilon \left(1-\left(1-\frac{1}{\lambda}\right)^{2k}
\right).
$$
Note that $(1-1/\lambda)^{2k} \le (1-1/\lambda)^{2r\lambda} \le
1/{e^{2r}}$. Therefore
\begin{equation}\label{bound_on_q}
q \ge s \varepsilon \left(1-\frac{1}{e^{2r}}\right) = sc,
\end{equation}
where $c=\varepsilon \left(1-\frac{1}{e^{2r}}\right)$. In what
follows we shall use the fact that conditions~(\ref{N_condition})
and (\ref{bound_on_q}) imply
\begin{equation} \label{useful}
\lambda \ge \frac{2}{s \varepsilon \left(1-1/e^{2r}\right)} \ge
2/q.
\end{equation}
To bound probability~$p$ from below, we first note that for any $z
\in [0,1]$ holds
\begin{equation}\label{simple} 1-\frac{z}{e} \ge e^{-z}.
\end{equation}
Assume $z=e^{-q \lambda/2+1}$. Then in view of
inequality~(\ref{useful}), $z\le 1$, and consequently,
\begin{equation}\label{lower_bound_on_p}
p \ge \exp\left\{-e^{1-q \lambda/2}\right\} \ge
\exp\left\{-e^{1-sc\lambda/2}\right\}.
\end{equation}
Now the right-hand side expression from~(\ref{lower_bound_on_p})
may be used as a lower
bound~$\ell=\exp\left\{-e^{1-sc\lambda/2}\right\}$.

Let us now consider a sequence of populations~$X^0,X^1,\dots$.
Note that $\ell^m$ is a lower bound for the probability to reach a
local optimum in a series of at most~$m$ iterations, where that at
each iteration the best found solution is improved, until a local
optimum is found. Indeed,
suppose~$A_{t}=E_1^{t}+\dots+E_{\lambda/2}^{t}, \ t=1,2,\dots$.
Then
\begin{equation}\label{cond_probab}
{\bf P}\{A_1\& \dots \& A_m\} = {\bf P}\{A_1\} \prod_{t=1}^{m-1}
{\bf P}\{A_{t+1}|A_1 \& \dots \& A_{t}\} \ge \ell^m.
\end{equation}

In view of condition~(\ref{N_condition}), we find a lower bound
for the probability to reach a local optimum in a sequence of at
most~$m$ iterations where the best found solution is improved in
each iteration:
$$
\ell^m = \exp\left\{-m e^{1-sc\lambda/2}\right\} \ge
 \exp\left\{-m e^{-\ln m}\right\}=1/e.
$$
$\Mybox$\\

Many well-known NP~optimization problems, such as the Maximum
Satisfiability Problem and the Maximum Cut Problem have a set of
feasible solutions equal to the whole search
space~$\{0,1\}^{n(I)}$. The following proposition applies to the
problems with such property.

\begin{proposition}\label{proposition_GA_LS1}
If \ ${\rm Sol(I)}=\{0,1\}^{n(I)}$ for all $I \in {\rm Inst}$ and
the conditions of Lemma~\ref{lemma_GA_LS} hold, then a local
optimum is reached in at most~$em$ iterations of the~$\mathcal GA$
on average.
\end{proposition}
{\bf Proof.} Consider a sequence of series of the $\mathcal GA$
iterations, where the length of each series is~$m$ iterations.
Suppose, $D_i, \ i=1,2,\dots,$ denotes an event of absence of
local optima in the population throughout the $i$-th series. The
probability of each event~$D_i, \ i=1,2,\dots,$ is at most
$\mu=1-1/e$ according to Lemma~\ref{lemma_GA_LS}. Analogously to
the bound~(\ref{cond_probab}) we obtain the inequality ${\bf
P}\{D_1\& \dots \& D_i\} \le \mu^i.$

Let~$\eta$ denote the random variable, equal to the number of the
first run where a local optimum was obtained. By the properties of
expectation (see e.g.~\cite{Gnedenko}),
$$ E[\eta] = \sum_{i=0}^{\infty} {\bf P}\{\eta > i\} =
1+\sum_{i=1}^{\infty} {\bf P}\{D_1\& \dots \& D_i\} \le 1
+\sum_{i=1}^{\infty} \mu^i = e.
$$
Consequently, the average number of iterations until a local
optimum is first obtained is at most~$em$.
$\Mybox$\\

Suppose the termination condition in the iterated $\mathcal GA$ is
{$t>t_{\max}=m$}. Then execution of this algorithm may be viewed
as a sequence of independent runs of the $\mathcal GA$, where the
length of each run is~$m$ iterations.

Let~$\lceil \cdot \rceil$ denote rounding up. In conditions of
Lemma~\ref{lemma_GA_LS}, given the parameters
\begin{equation}\label{ga_settings}
\lambda = 2 \left\lceil \frac{1+ \ln m}{s\varepsilon(1-1/e^{2r})}
\right\rceil, \quad k = \lceil r\lambda \rceil,
\end{equation}
the probability that $\mathcal GA$ finds a local optimum during
the first~$m$ iterations is~$1/e$.
So the total number of populations computed in the iterated
$\mathcal GA$ until it first visits a local optimum is at
most~$em$.

The operators~$\mbox{Mut}$ and $\mbox{Cross}$ are supposed to be
efficiently computable and the tournament selection
requires~$O(\lambda)$ time. Therefore the time complexity of
computing a pair of offspring in the $\mathcal GA$ is polynomially
bounded and the following theorem holds.

\begin{theorem}\label{GA_LS}
If problem~$\Pi=({\rm Inst},\mbox{\rm Sol}(I),f_I)$ and the
function~$s^{-1}(I)$ are polynomially bounded and population~$X^0$
contains a feasible solution at every run, then the
iterated~$\mathcal GA$ with suitable choice of parameters first
visits a local optimum on average in polynomially bounded time.
\end{theorem}

Note that a slight modification of the proof of Theorem~4
from~\cite{Lehre2011} yields the result of the above theorem in
the case when the crossover operator is not used. The proof
in~\cite{Lehre2011}, however, is based on a more complex method of
drift analysis.

Often the neighborhood mappings for NP~optimization problems are
polynomially bounded, i.e. the cardinality $|\mathcal{N}({\bf
x})|,$ ${\bf x} \in {\rm \mbox{\rm Sol}}$ is a polynomially
bounded value~\cite{RT87}. In such cases there exists a mutation
operator~$\mbox{Mut}(\x)$ that generates a uniform distribution
over the set~$\mathcal{N}({\bf x})$, and the condition on
function~$s^{-1}(I)$ in Theorem~\ref{GA_LS} is satisfied.\\

Let $\delta({\bf x},{\bf y})$ denote the Hamming distance between
${\bf x}$ and ${\bf y}$.

\begin{definition}\label{k_bounded} {\rm \cite{AP95}}
Suppose $\Pi$ is an NP~optimization problem. A neighborhood
mapping~$\mathcal{N}$ is called $K$-bounded, if for any ${\bf x}
\in {\rm \mbox{\rm Sol}}$ and ${\bf y} \in \mathcal{N}({\bf x})$
holds $\delta({\bf x},{\bf y}) \le K$, where $K$ is a constant.
\end{definition}

The bitwise mutation operator~$\mbox{Mut}^*$ with probability
$P_{\rm m}^{\delta({\bf x},{\bf y})}(1-P_{\rm m})^{n-\delta({\bf
x},{\bf y})}$ outputs a string~${\bf y}$, given a string~${\bf
x}$. Note that probability~$P_{\rm m}^j(1-P_{\rm m})^{n-j}$, as a
function of~$P_{\rm m}$, \ $P_{\rm m} \in [0,1]$, attains its
minimum at $P_{\rm m}=j/n$. The following proposition gives a
lower bound for the probability ${\bf P}\{{\mbox{Mut}^*({\bf
x})={\bf y}}\}$, which is valid for any ${\bf y} \in
\mathcal{N}({\bf x})$, assuming that~$P_{\rm m}=K/n$.

\begin{proposition}\label{optimal_bound}
Suppose the neighborhood mapping~$\mathcal{N}$ is $K$-bounded,
$K\le n/2$ and $P_{\rm m}=K/n$. Then for any ${\bf x} \in {\rm
\mbox{\rm Sol}}$ and any ${\bf y} \in \mathcal{N}({\bf x})$ holds
$$
{\bf P}\{\mbox{Mut}^*({\bf x})={\bf y}\} \ge
\left(\frac{K}{en}\right)^K.
$$
\end{proposition}

{\bf Proof.} For any ${\bf x} \in {\rm \mbox{\rm Sol}}$ and ${\bf
y} \in \mathcal{N}({\bf x})$ we have
$$
{\bf P}\{\mbox{Mut}^*({\bf x})={\bf y}\}
=\left(\frac{K}{n}\right)^{\delta({\bf x},{\bf
y})}\left(1-\frac{K}{n}\right)^{n-\delta({\bf x},{\bf y})} \ge
 \left(\frac{K}{n}\right)^K
 \left(1-\frac{K}{n}\right)^{n-K},
$$
since $P_{\rm m} =K/n\le 1/2$. Now $\frac{\partial}{
\partial n} (1-K/n)^{n-K} <0$ for $n>K$, and besides that, $(1-K/n)^{n-K} \to 1/e^K$
as~$n\to \infty$. Therefore $(1-K/n)^{n-K} \ge 1/e^K$, which
implies the required inequality. $\Mybox$\\

\section{Analysis of Guaranteed Local Optima Problems}
\label{sec:GLO}

In this section Theorem~\ref{GA_LS} is used to estimate the GA
capacity of finding the solutions with guaranteed approximation
ratio.

\begin{definition}\label{GLO} {\rm \cite{AP95}}
A polynomially bounded NP~optimization problem~$\Pi$ belongs to
the class~{\rm GLO} of {\em Guaranteed Local Optima} problems, if
the following two conditions hold:

1) At least one feasible solution~${\bf y}_I \in {\rm \mbox{\rm
Sol}}$ is efficiently computable for every instance~$I \in {\rm
Inst}$;

2) A $K$-bounded neighborhood mapping~$\mathcal{N}_I$ exists, such
that for every instance~$I$, any local optimum of~$I$ with respect
to~$\mathcal{N}_I$ has a constant guaranteed approximation ratio.

\end{definition}

The class~{\rm GLO} contains such well-known {\rm NP}~optimization
problems as Maximum Staisfiablity, Maximum Cut and the following
problems on graphs with bounded vertex degree: Independent Set
Problem, Dominating Set Problem and Vertex Cover~\cite{AP95}.

If a problem~$\Pi$ belongs to GLO and $n/2 \ge K$ then in view of
Proposition~\ref{optimal_bound}, for any  ${\bf x} \in {\rm
\mbox{\rm Sol}}$ and ${\bf y} \in \mathcal{N}({\bf x})$, the
bitwise mutation operator with~$P_{\rm m}=K/n$ satisfies the
condition ${\bf P}\{\mbox{Mut}^*({\bf x})={\bf y}\} \ge 1/{\rm
poly}(|I|)$, where poly is some polynomial. If $n/2<K<n$, then
probability ${\bf P}\{\mbox{Mut}^*({\bf x})={\bf y}\}$ is bounded
from below by a positive constant. Therefore, Theorem~\ref{GA_LS}
implies the following

\begin{corollary}\label{GA_GLO}
If $\Pi\in {\rm GLO}$ and population~$X^0$ at every run contains a
feasible solution, then given suitable values of parameters, the
iterated~$\mathcal GA$ with bitwise mutation first visits a
solution with a constant guaranteed approximation ratio in
polynomially bounded time on average.
\end{corollary}


\section*{\large{\textbf{Conclusion}}}

The obtained results indicate that if a local optimum is
efficiently computable by the local search method, it is also
computable in expected polynomial time by the iterated~GA with
tournament selection. The same applies to the GA without restarts,
if the set of feasible solutions is the whole search space.
Besides that, given suitable parameters, the iterated GA with
tournament selection and bitwise mutation approximates any problem
from GLO class within a constant ratio in polynomial time on
average.


\section{Acknowledgements} Supported by Russian Foundation
for Basic Research grants 12-01-00122 and 13-01-00862.

\end{document}